\documentclass[a4paper,twoside]{article}

\usepackage{epsfig}
\usepackage{subcaption}
\usepackage{calc}
\usepackage{amssymb}
\usepackage{amstext}
\usepackage{amsmath}
\usepackage{amsthm}
\usepackage{multicol}
\usepackage{pslatex}
\usepackage{apalike}
\usepackage{algorithm2e}
\usepackage{array}
\usepackage{graphicx}
\usepackage{multirow}
\usepackage{booktabs}
\usepackage{colortbl}
\usepackage[bottom]{footmisc}

\usepackage{SCITEPRESS}     

\usepackage{changes}
\usepackage{xcolor}
\begin{document}

\title{3D Gaussian Splatting with Fisheye Images: Field of View Analysis and Depth-Based Initialization}

\author{\authorname{Ulas Gunes\sup{1}, Matias Turkulainen \sup{2},  Mikhail Silaev\sup{1}, Juho Kannala \sup{2,3}, Esa Rahtu \sup{1}}
\affiliation{\sup{1}Department of Computing Sciences, Tampere University, Korkeakoulunkatu 7, Tampere, Finland}
\affiliation{\sup{2}Department of Computer Science, Aalto University, Otakaari, Espoo, Finland}
\affiliation{\sup{2}Department of Computer Science and Engineering, University of Oulu, Pentti Kaiteran katu 1, Oulu, Finland}
\email{\{ulas.gunes, mikhail.silaev, esa.rahtu\}@tuni.fi, \{matias.turkulainen, juho.kannala\}@aalto.fi}
}


\abstract{We present the first evaluation of 3D Gaussian Splatting methods on real fisheye imagery with fields of view above 180\textdegree{}. Our study evaluates Fisheye-GS \cite{liao2024fisheyegslightweightextensiblegaussian} and 3DGUT \cite{wu20253dgut} on indoor and outdoor scenes captured with 200\textdegree{} fisheye cameras, with the aim of assessing the practicality of wide-angle reconstruction under severe distortion. By comparing reconstructions at 200\textdegree{}, 160\textdegree{}, and 120\textdegree{} field-of-view, we show that both methods achieve their best results at 160\textdegree{}, which balances scene coverage with image quality, while distortion at 200\textdegree{} degrades performance.
To address the common failure of Structure-from-Motion (SfM) initialization at such wide angles, we introduce a depth-based alternative using UniK3D (Universal Camera Monocular 3D Estimation) \cite{piccinelli2025unik3d}. This represents the first application of UniK3D to fisheye imagery beyond 200\textdegree{}, despite the model not being trained on such data. With the number of predicted points controlled to match SfM for fairness, UniK3D produces geometrically accurate reconstructions that rival or surpass SfM, even in challenging scenes with fog, glare, or open sky. These results demonstrate the feasibility of fisheye-based 3D Gaussian Splatting and provides a benchmark for future research on wide-angle reconstruction from sparse and distorted inputs.
}

\onecolumn \maketitle \normalsize \setcounter{footnote}{0} \vfill
\section{\uppercase{Introduction}}
\label{sec:introduction}

3D Gaussian Splatting (3DGS) \cite{kerbl3Dgaussians} has become a foundational technique for high-quality 3D scene reconstruction and real-time image-based rendering, with wide adoption in both academia and industry. While many extensions improve rendering, editing, and optimization, most operate under narrow field-of-view (FoV) perspective cameras due to their simple geometry and mature calibration pipelines.

Fisheye cameras, by contrast, provide ultra-wide FoVs that capture larger portions of a scene with fewer images. This makes them attractive for tasks in autonomous driving, robotics, and VR/AR, where reducing sensor count and capture time is crucial. Fewer viewpoints can also lower the computational and memory requirements of reconstruction. However, fisheye-based 3D reconstruction remains relatively unexplored due to the nonlinear projection and strong radial distortion characteristics of such lenses.

Recent approaches such as Fisheye-GS \cite{liao2024fisheyegslightweightextensiblegaussian} and 3DGUT \cite{wu20253dgut} extend 3DGS to fisheye images by modifying projection and rendering to support non-linear camera models. Although these methods are promising, their real-world behavior across different fields of view and diverse scene types has not been comprehensively evaluated.

Initialization poses an additional challenge. Standard Structure-from-Motion (SfM) pipelines assume perspective projection and struggle with distortion-heavy fisheye images. Monocular 3D estimation offers a lightweight alternative, yet most models generalize poorly to wide-angle data. UniK3D \cite{piccinelli2025unik3d} supports arbitrary intrinsics, including fisheyes, but it is trained mainly on synthetic wide-FoV data, leaving its performance on real fisheye imagery largely untested.

In this work, we explore whether monocular 3D estimation can reliably replace SfM for initializing fisheye-based 3DGS pipelines. Specifically, we evaluate how well UniK3D predictions can substitute SfM point clouds when used with splatting pipelines designed for non-linear projection models. Our goal is to assess both compatibility and practical reconstruction quality across a diverse set of real-world scenes.

\paragraph{Contributions} We present the first evaluation of the fisheye-adapted 3D Gaussian Splatting methods, Fisheye-GS and 3DGUT, on real images with fields of view exceeding 180\textdegree{} in both indoor and outdoor scenes. We provide the first empirical analysis of UniK3D on real fisheye data with ultra-wide fields of view and test its effectiveness as an alternative to SfM-based initialization. We show that monocular estimates from only 2–3 fisheye images per scene can be converted into point clouds that yield reconstruction quality comparable to SfM. We further evaluate how reducing the field of view (FoV) affects reconstruction quality and analyze the trade-off between peripheral distortion and scene coverage. Finally, we align monocular point clouds to the COLMAP coordinate frame, for compatibility with existing 3D Gaussian Splatting pipelines.

\section{\uppercase{Related Work}}

\subsection{Novel View Synthesis and 3D Gaussian Splatting}

Novel view synthesis aims to generate realistic images from unseen viewpoints. Neural Radiance Fields (NeRF) \cite{mildenhall2020nerf} model scenes as volumetric radiance fields and achieve high fidelity, but suffer from slow training and inference due to ray marching. While non-pinhole camera models such as fisheye cameras can be incorporated into Radiance Field based works \cite{smerf,zipnerf}, large-scale evaluations remain limited.

3D Gaussian Splatting (3DGS) \cite{kerbl3Dgaussians} addresses NeRF’s computational cost by representing scenes with anisotropic Gaussians rasterized directly in image space, enabling real-time rendering while preserving photorealism. Typical pipelines assume narrow-FoV perspective images and rely on SfM-based initialization (commonly COLMAP \cite{schoenberger2016sfm}). Although COLMAP supports fisheye models \cite{opencv_library}, parameter estimation degrades under strong distortion or very wide FoV, often requiring pre-calibrated intrinsics. This restricts scalability when dealing with uncalibrated fisheye data.

\subsection{Fisheye Extensions of 3D Gaussian Splatting}

Recent work has begun adapting 3DGS to non-linear projection models. Several methods propose modified projection functions or differentiable distortion handling \cite{huang2024erroranalysis3dgaussian,deng2025selfcalibratinggaussiansplattinglarge,jimaging10120330}, but results are typically limited or demonstrated on synthetic data.

Only a few approaches explicitly render circular fisheye images. Fisheye-GS \cite{liao2024fisheyegslightweightextensiblegaussian} replaces the perspective projection with an equidistant model and renders directly in the fisheye domain, assuming accurate calibration. 3DGUT \cite{wu20253dgut} replaces EWA splatting \cite{zwicker2002ewa} with an Unscented Transform \cite{UT}, enabling projection through arbitrary non-linear camera models and modeling additional effects such as reflections and rolling shutter. Both methods are summarized in the Appendix.

\subsection{Learning-Based Methods for Monocular Fisheye Depth Estimation}

Transformer-based monocular depth models such as Depth Anything \cite{depth_anything_v1,depth_anything_v2} and MoGe \cite{wang2024moge} perform well on perspective data but degrade under strong distortion. Fisheye-specific self-supervised methods \cite{9197319,DBLP:conf/ei-avm/KumarYMM21,kumar2023unrectdepthnetselfsupervisedmonoculardepth,zhao2024fisheyedepthrealscaleselfsupervised} address this by incorporating distortion-aware warping or unified projection models, but typically rely on fixed distortion assumptions and controlled hardware setups.

UniK3D \cite{piccinelli2025unik3d} is a recent transformer-based model supporting arbitrary intrinsics, including fisheye lenses, via a spherical scene representation and angular supervision. It predicts depth and ray directions without requiring predefined distortion models. We use UniK3D to obtain depth maps and rays from $200^\circ$ fisheye images and fuse them into dense point clouds that replace SfM initialization for 3DGS. Although not trained on real fisheye images with extreme FoV, we evaluate whether UniK3D’s predictions are sufficiently robust for reconstruction with Fisheye-GS and 3DGUT.

\section{\uppercase{Method}}
\begin{figure*}[hbt]
\centering
\makebox[0.19\textwidth][c]{\small GT}%
\hfill
\makebox[0.19\textwidth][c]{\small FGS - SfM}%
\hfill
\makebox[0.19\textwidth][c]{\small FGS - Depth}%
\hfill
\makebox[0.19\textwidth][c]{\small 3DGUT - SfM}%
\hfill
\makebox[0.19\textwidth][c]{\small 3DGUT - Depth}%

\vspace{0.1em}

\includegraphics[width=0.19\textwidth]{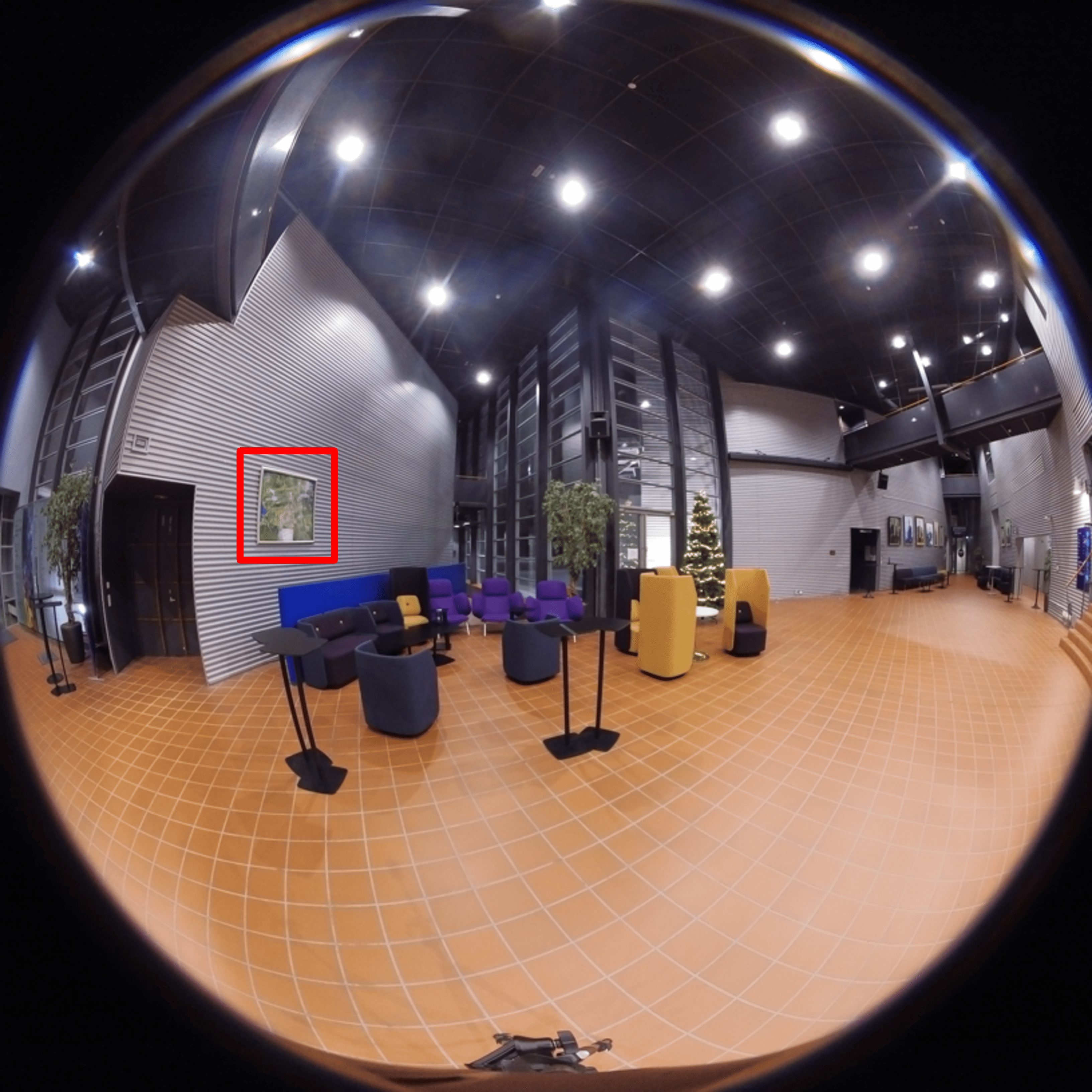}
\includegraphics[width=0.19\textwidth]{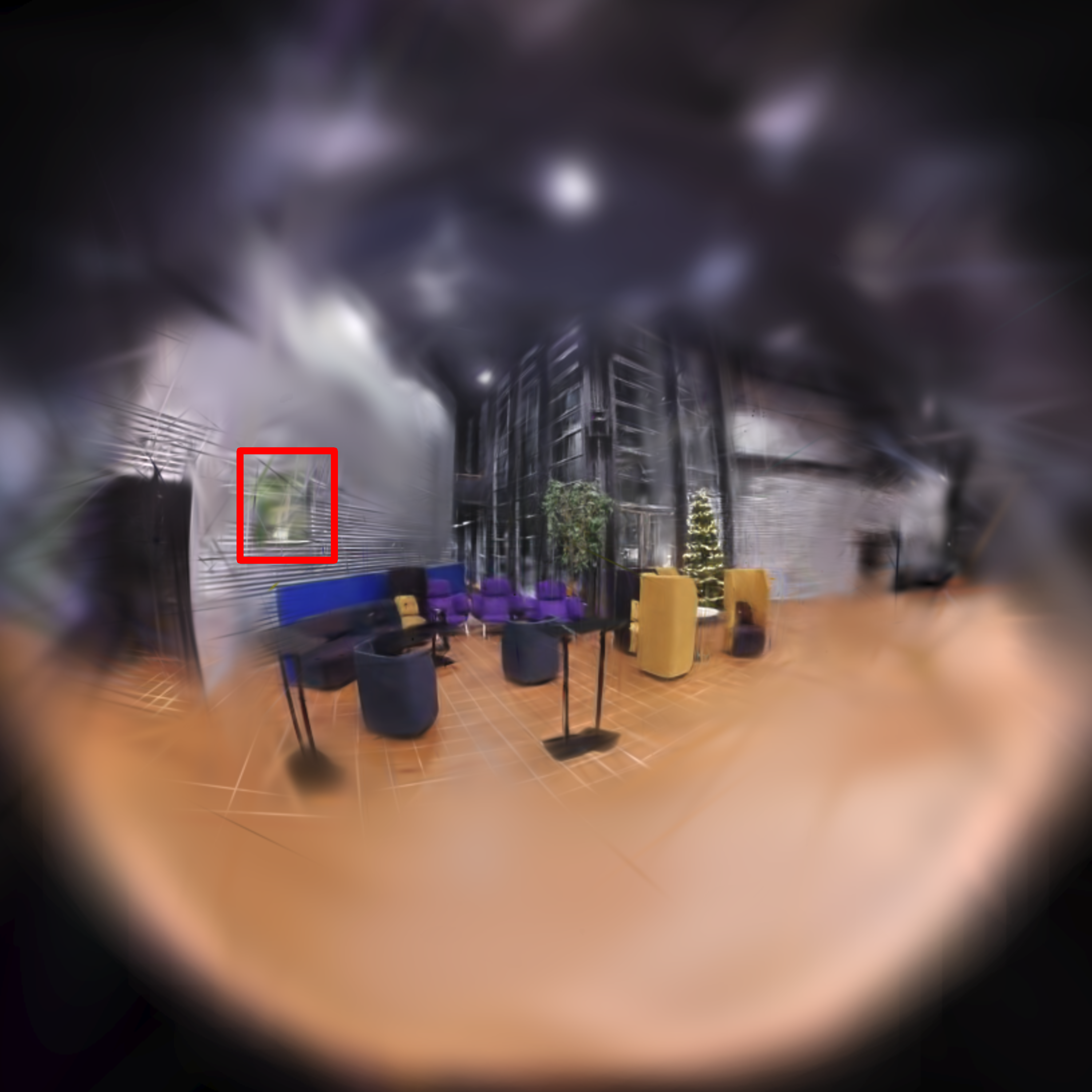}
\includegraphics[width=0.19\textwidth]{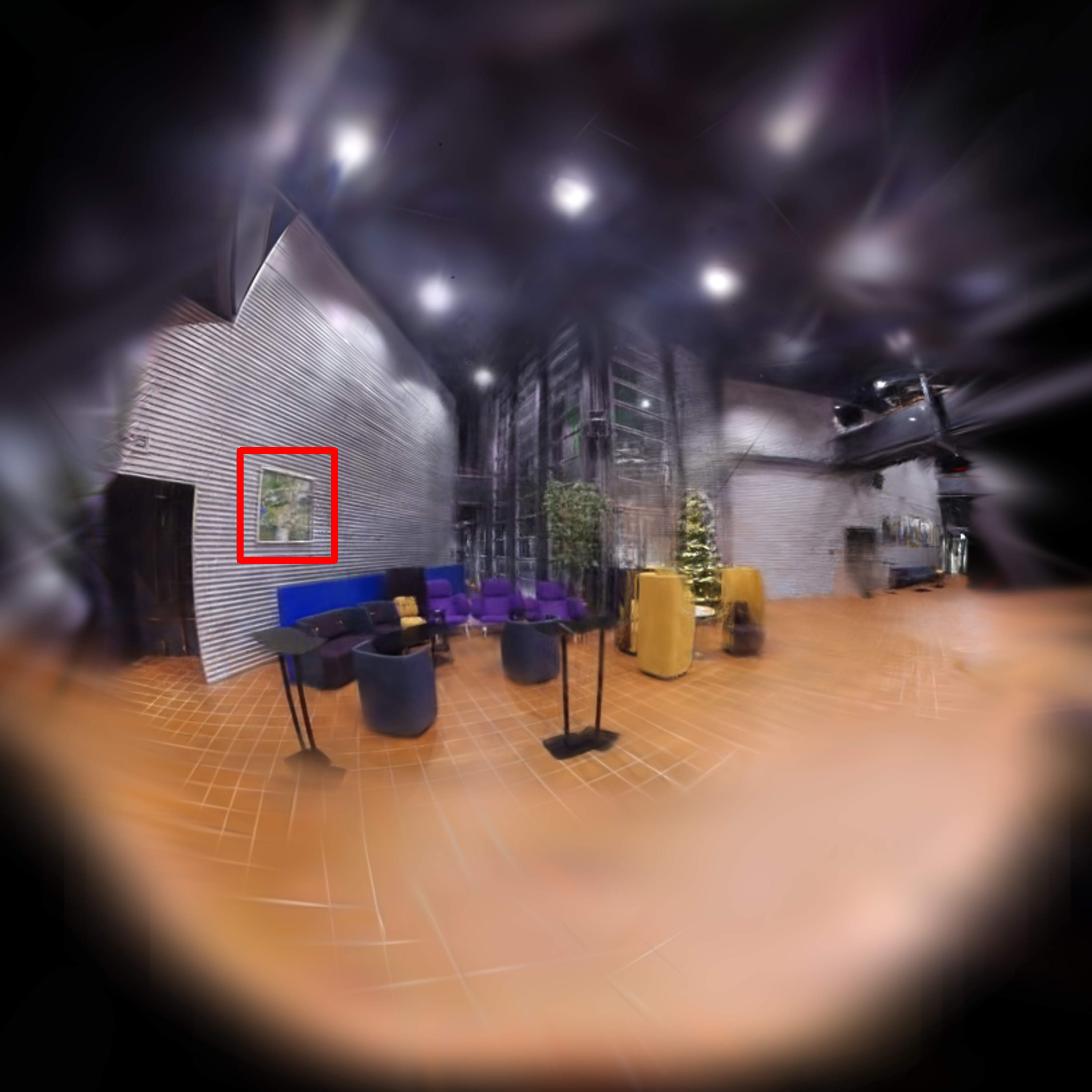}
\includegraphics[width=0.19\textwidth]{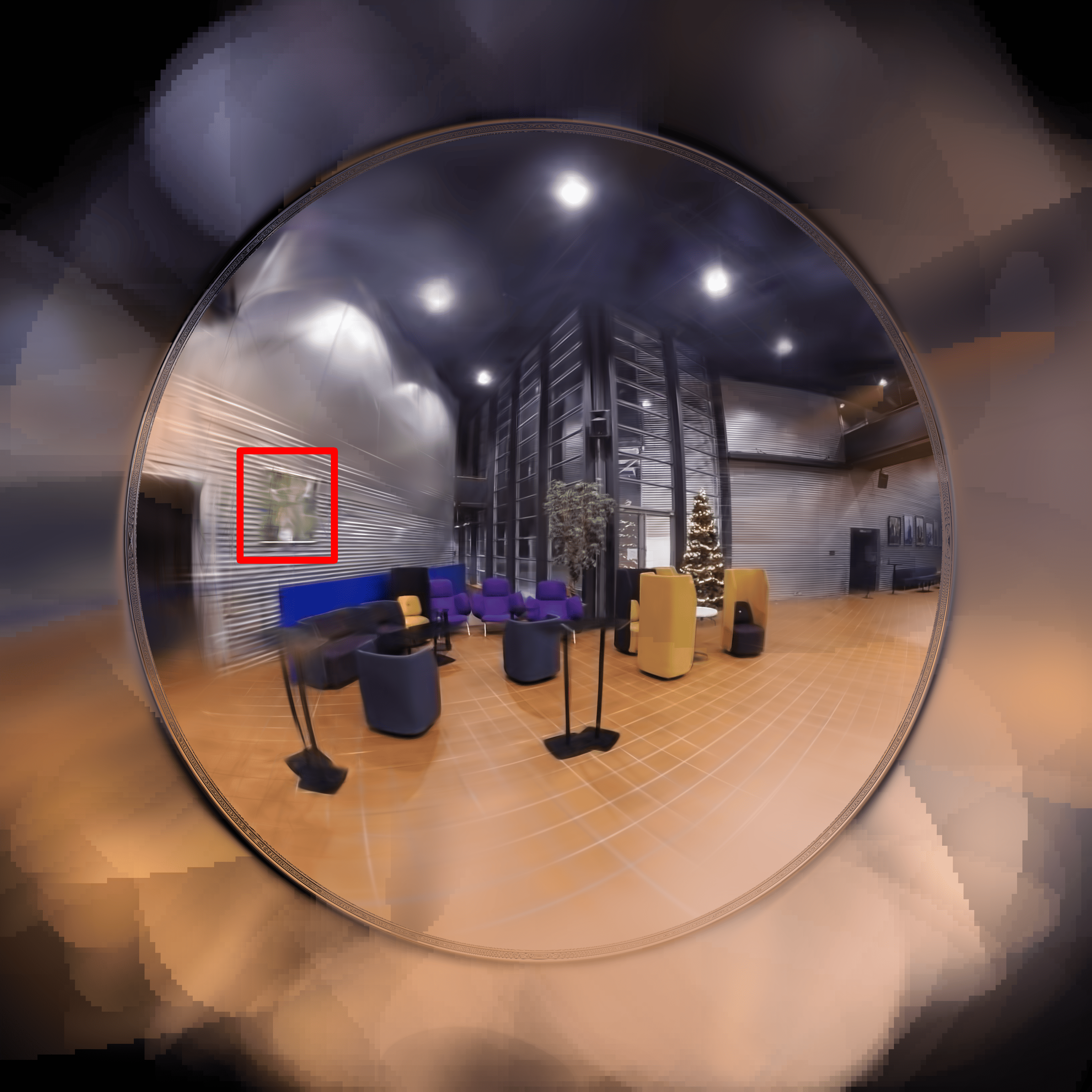}
\includegraphics[width=0.19\textwidth]{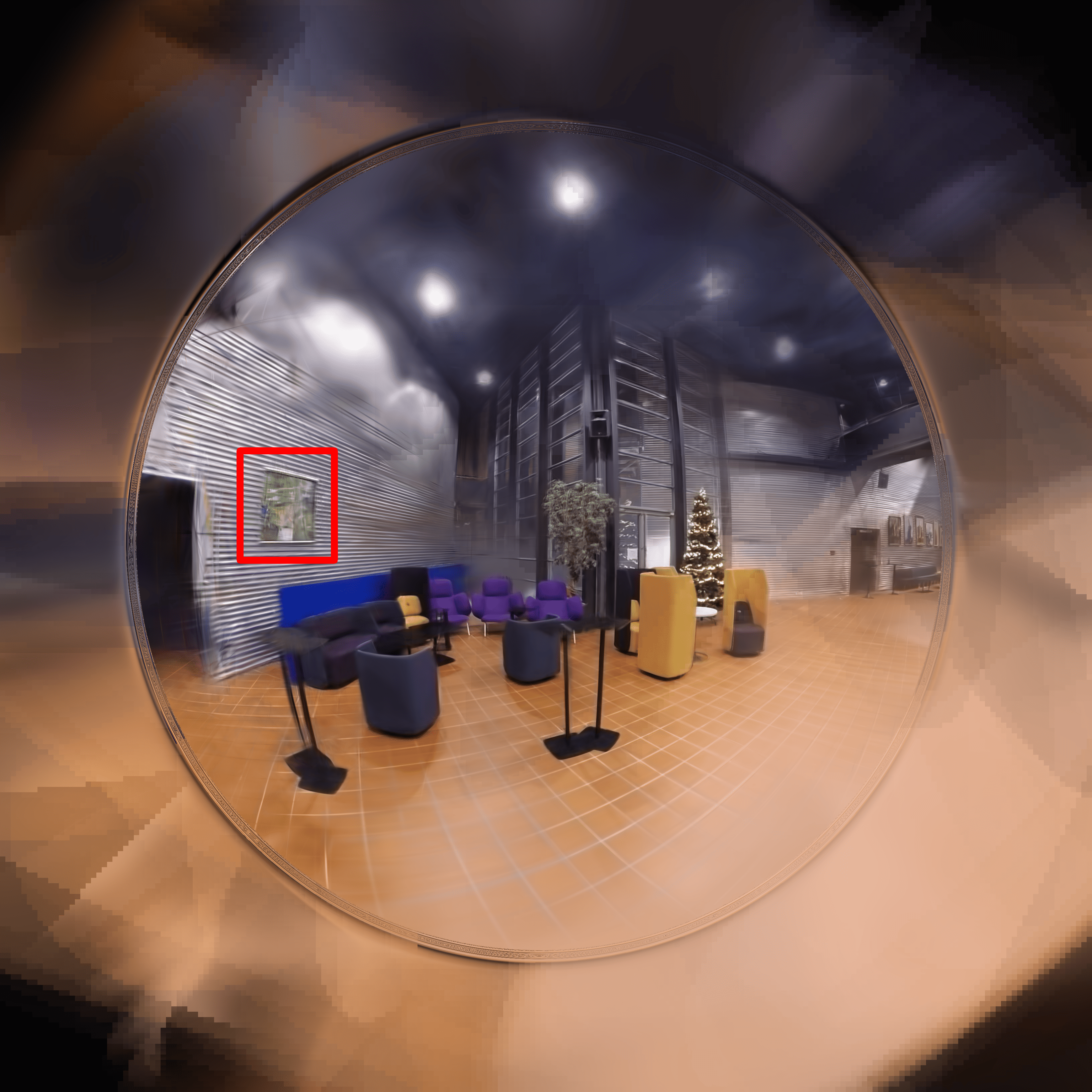}

\vspace{0.1em}

\includegraphics[width=0.19\textwidth]{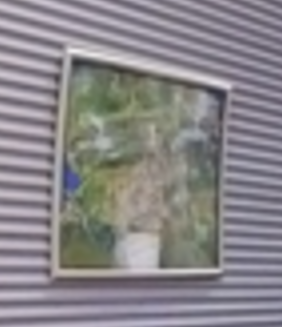}
\includegraphics[width=0.19\textwidth]{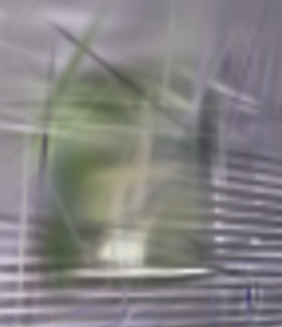}
\includegraphics[width=0.19\textwidth]{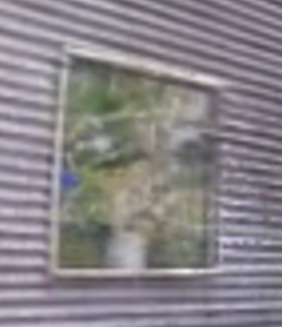}
\includegraphics[width=0.19\textwidth]{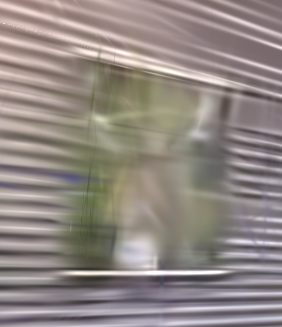}
\includegraphics[width=0.19\textwidth]{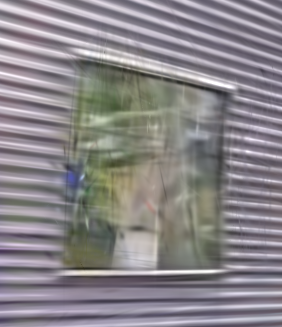}

\vspace{0.1em}

\includegraphics[width=0.19\textwidth]{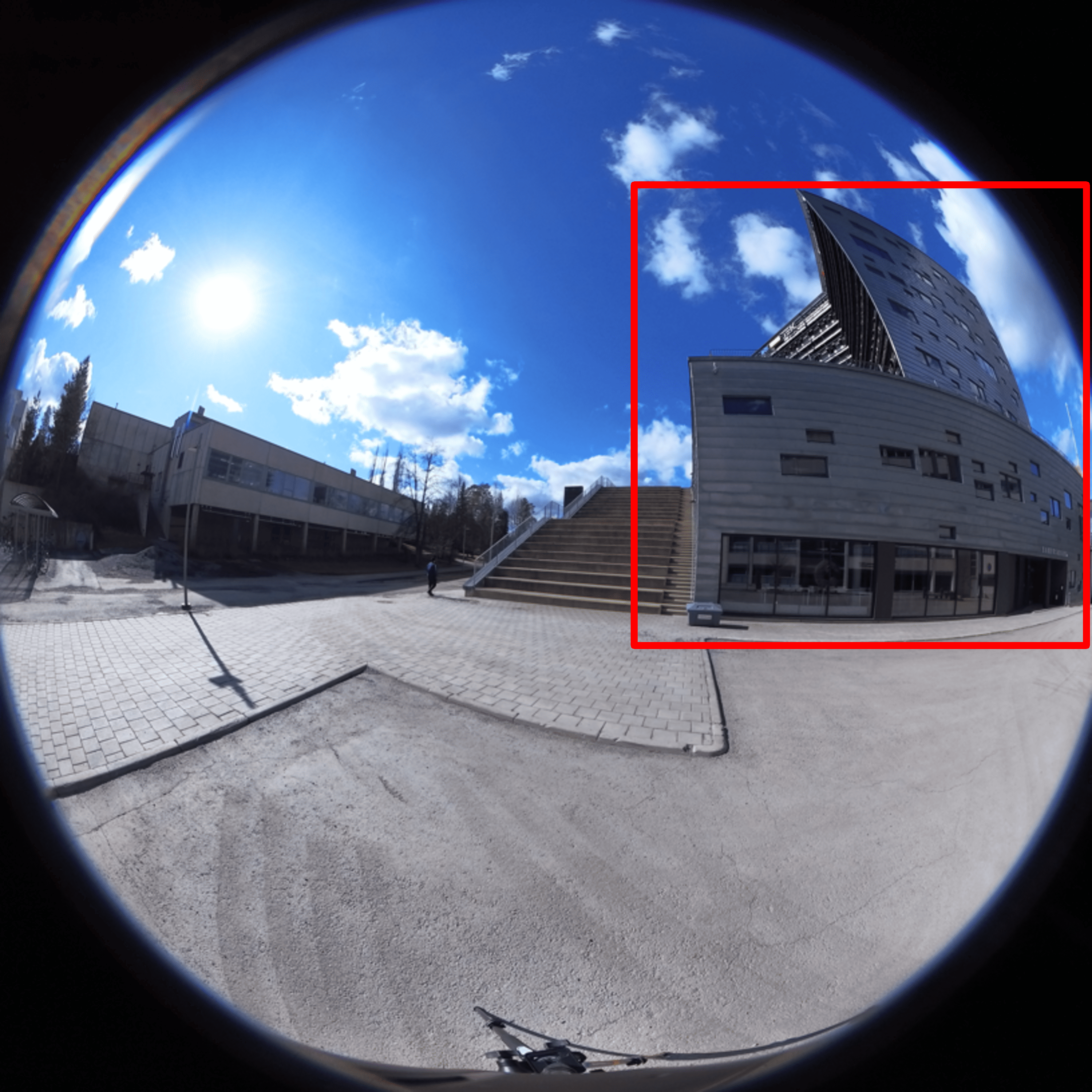}
\includegraphics[width=0.19\textwidth]{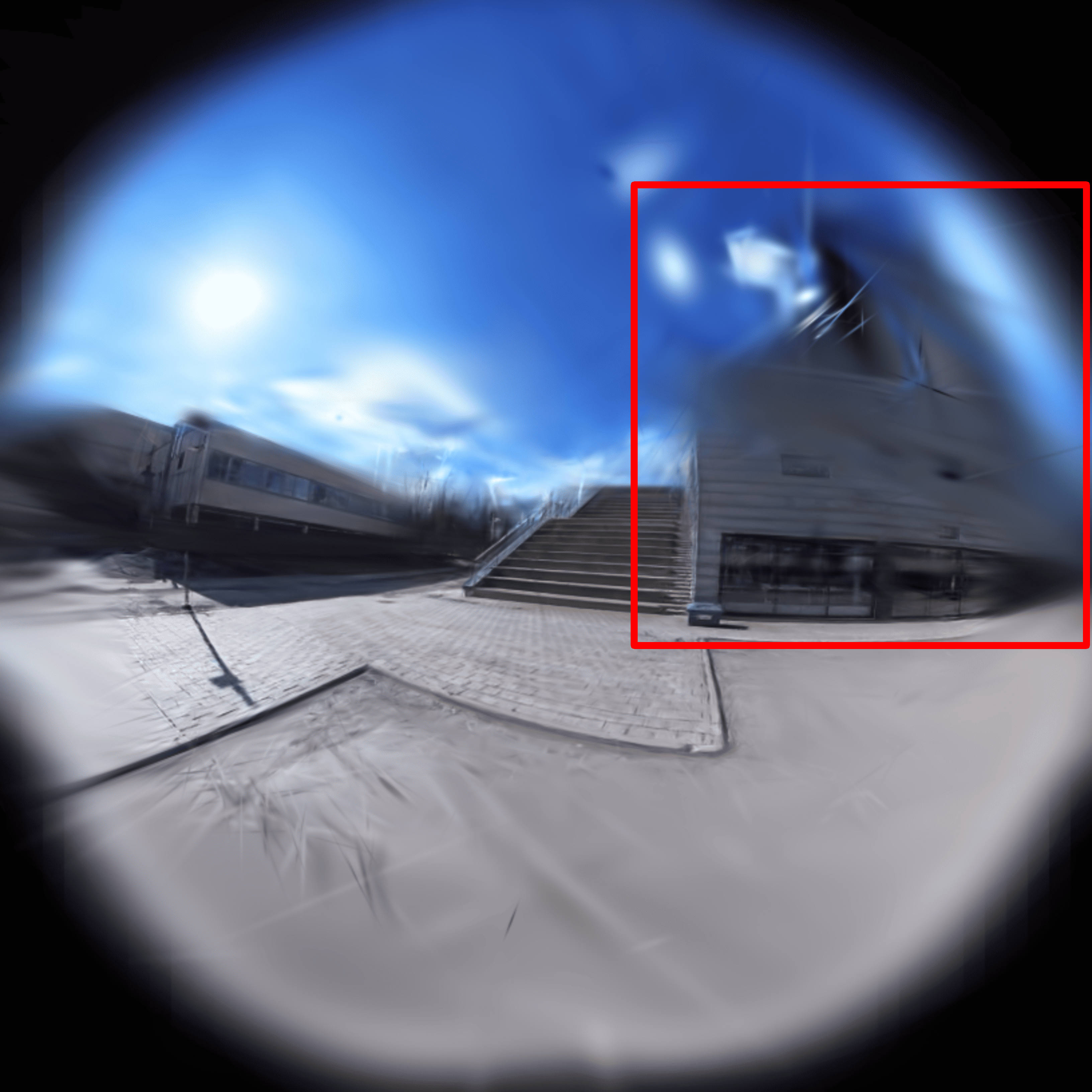}
\includegraphics[width=0.19\textwidth]{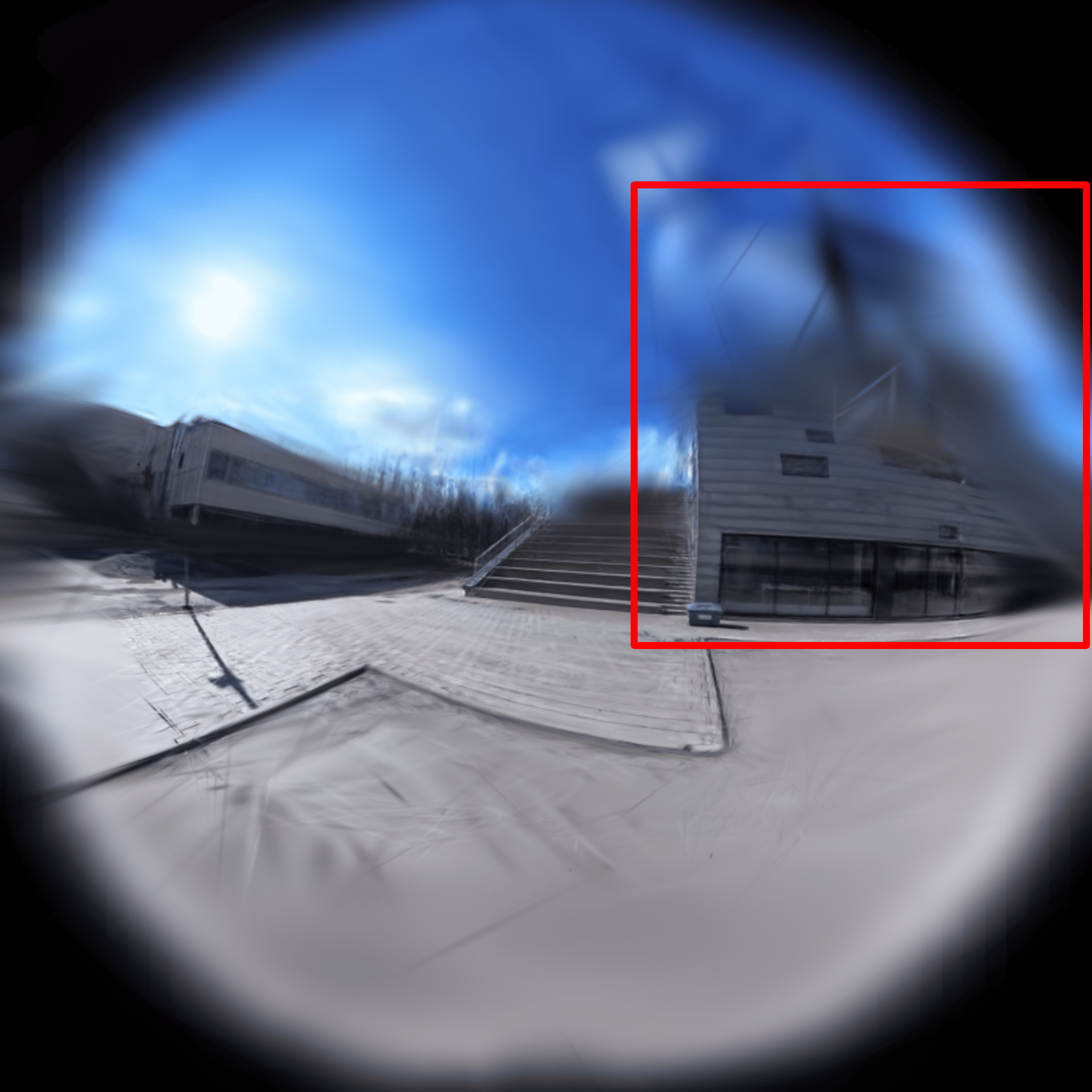}
\includegraphics[width=0.19\textwidth]{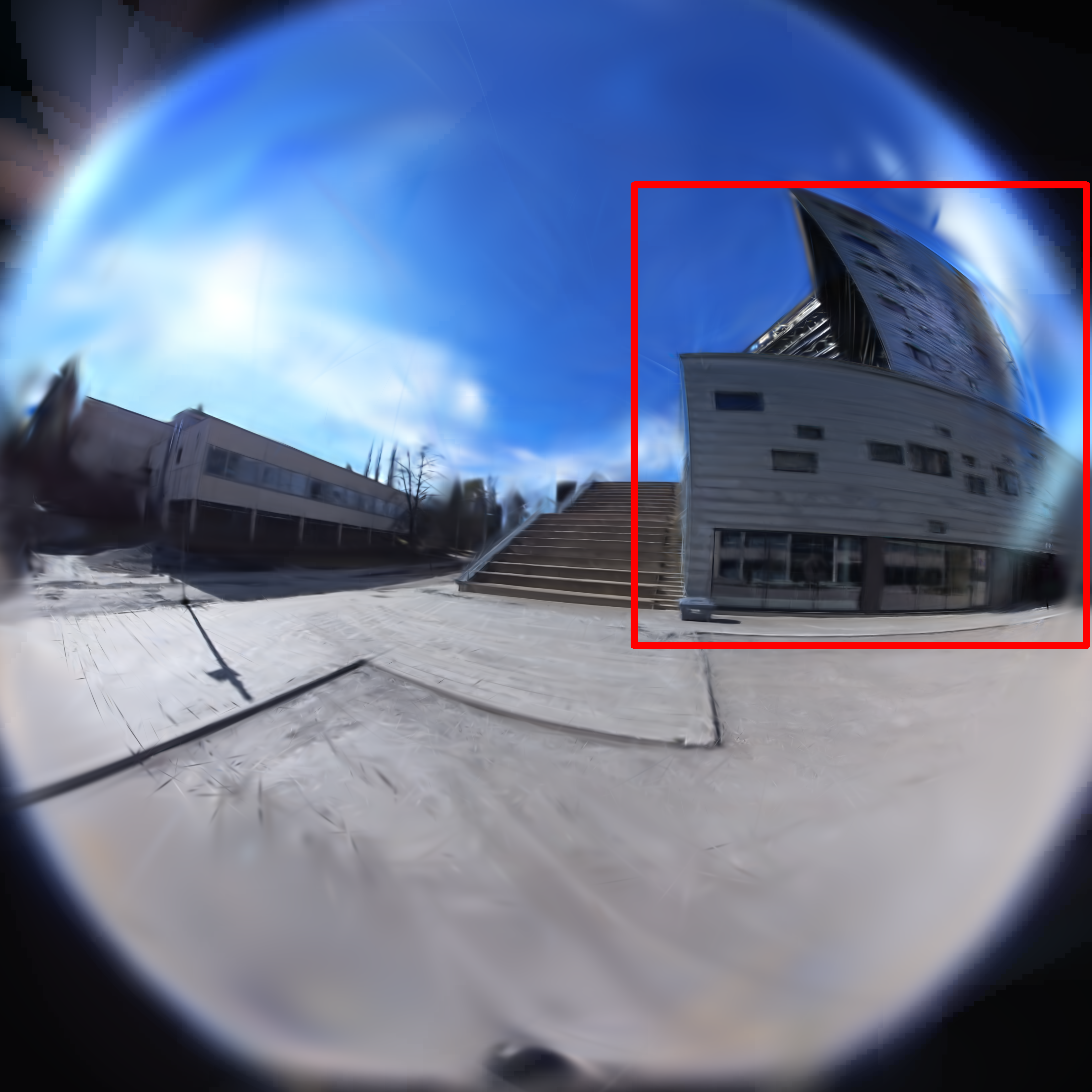}
\includegraphics[width=0.19\textwidth]{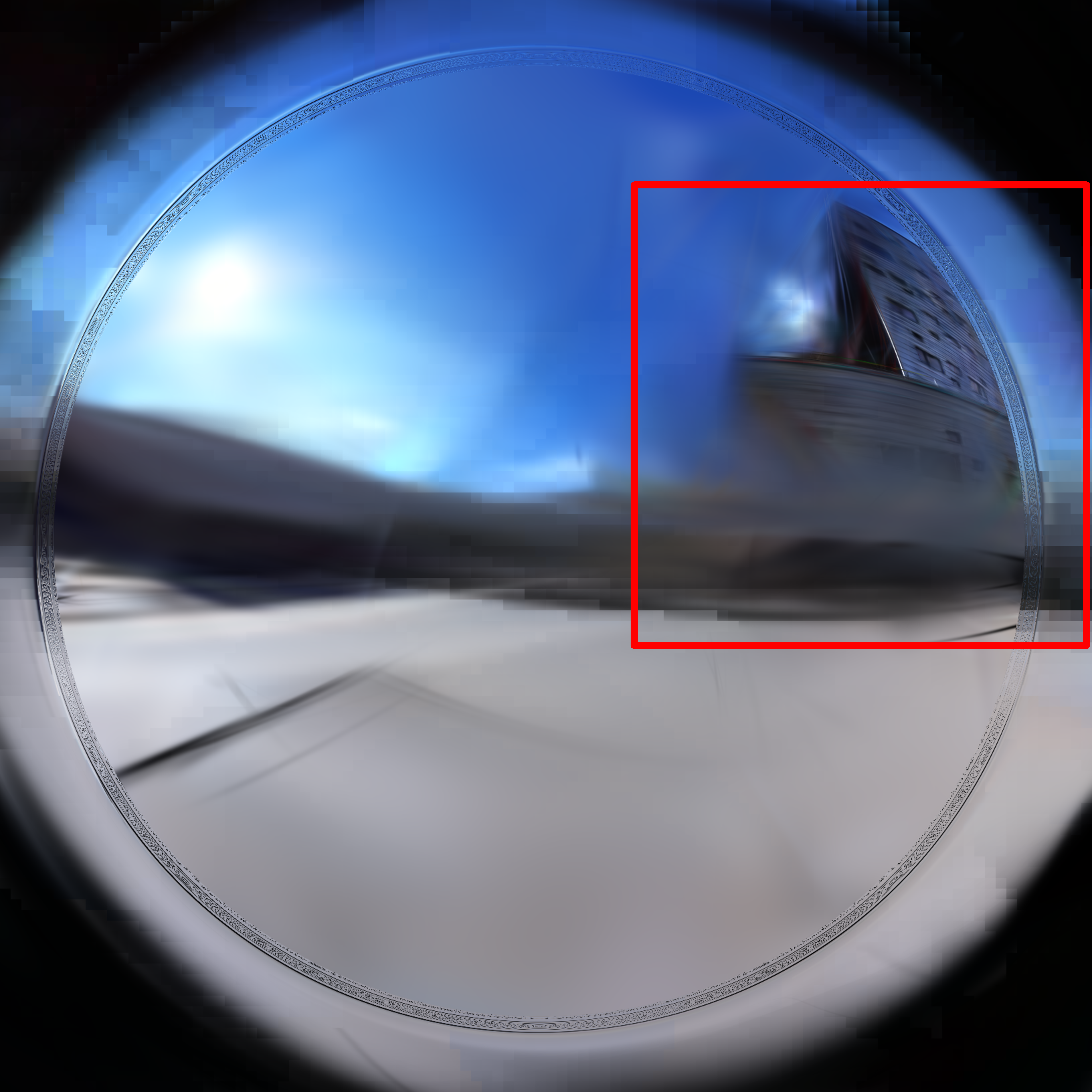}

\vspace{0.1em}

\includegraphics[width=0.19\textwidth]{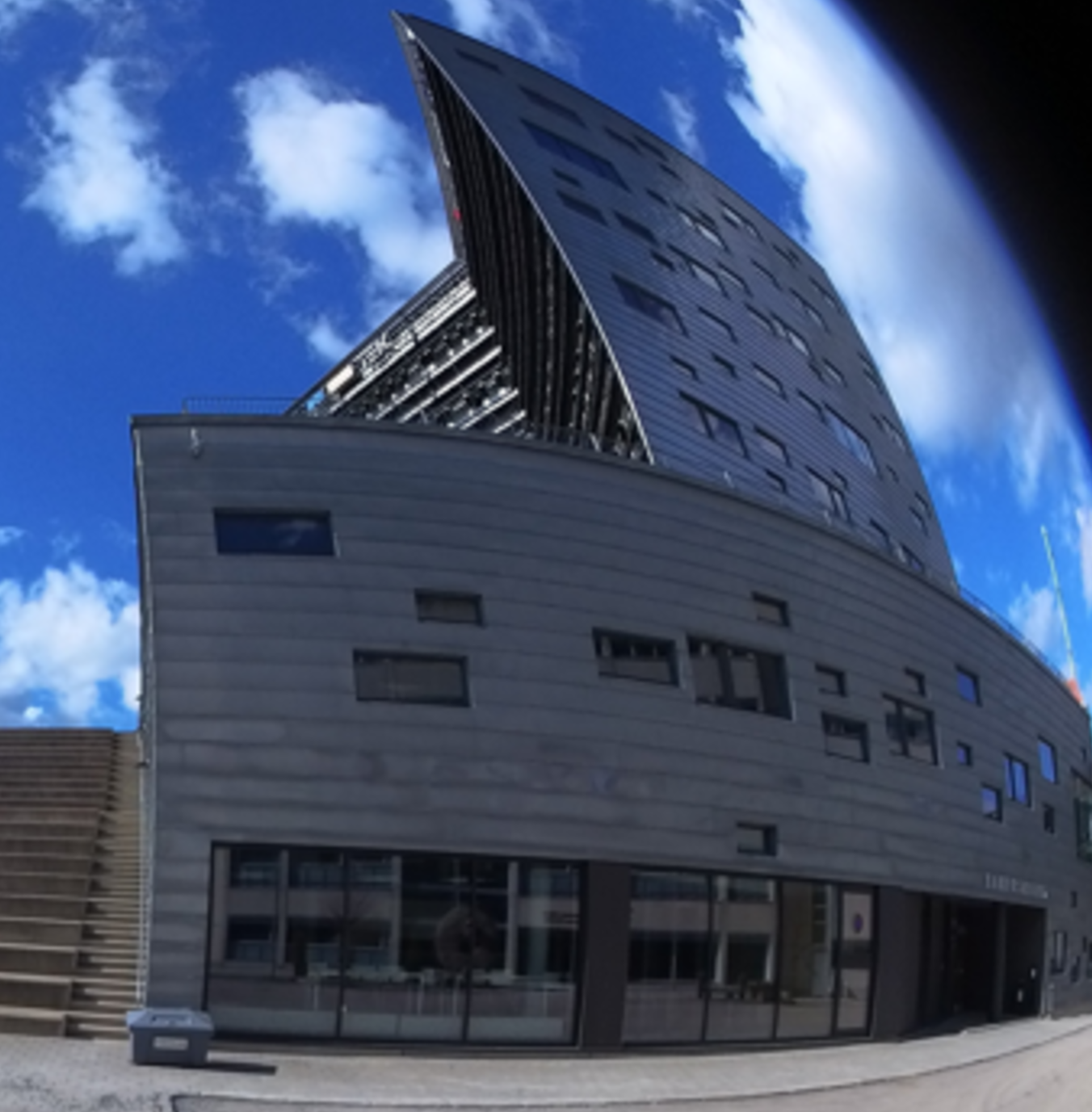}
\includegraphics[width=0.19\textwidth]{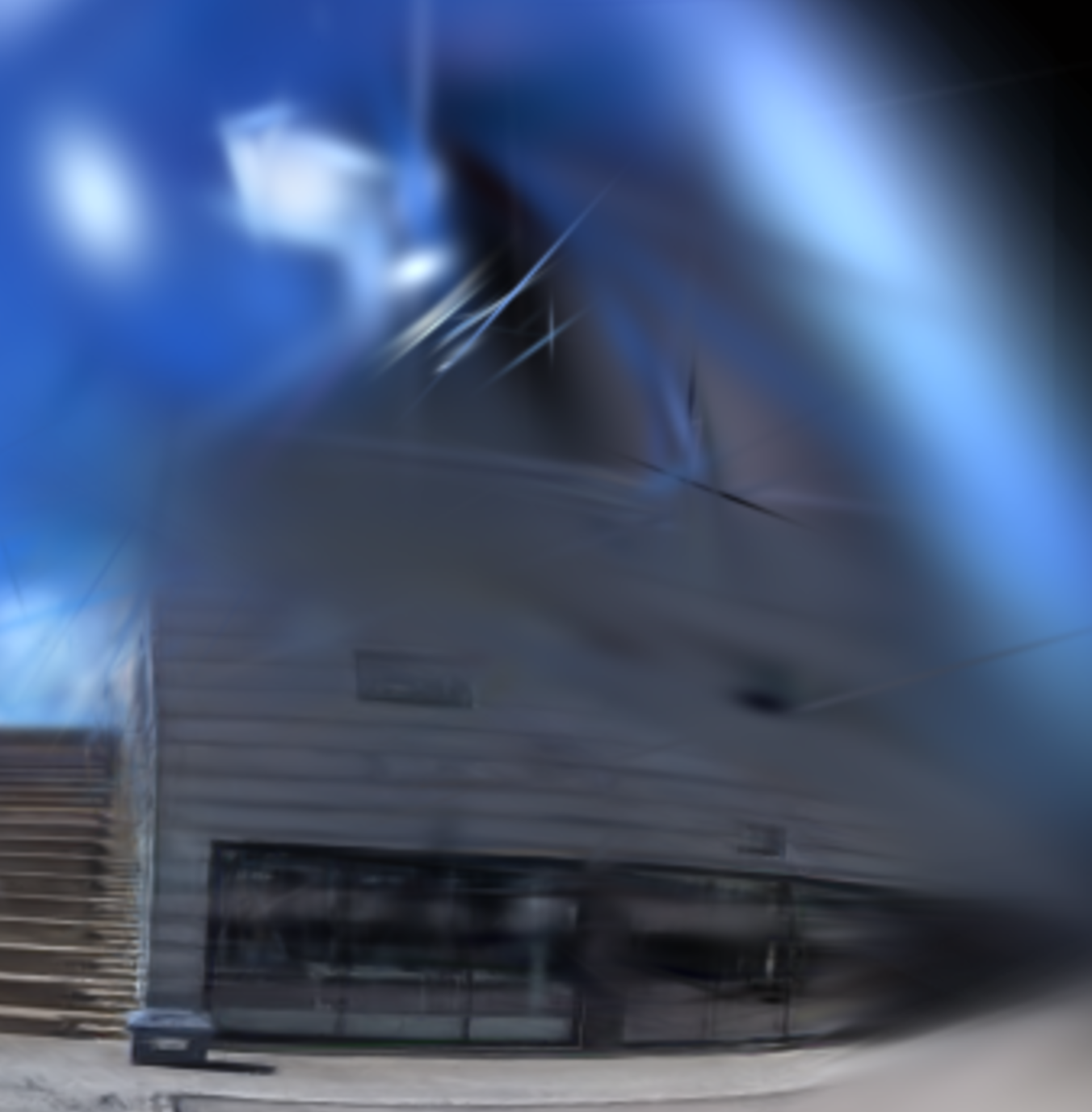}
\includegraphics[width=0.19\textwidth]{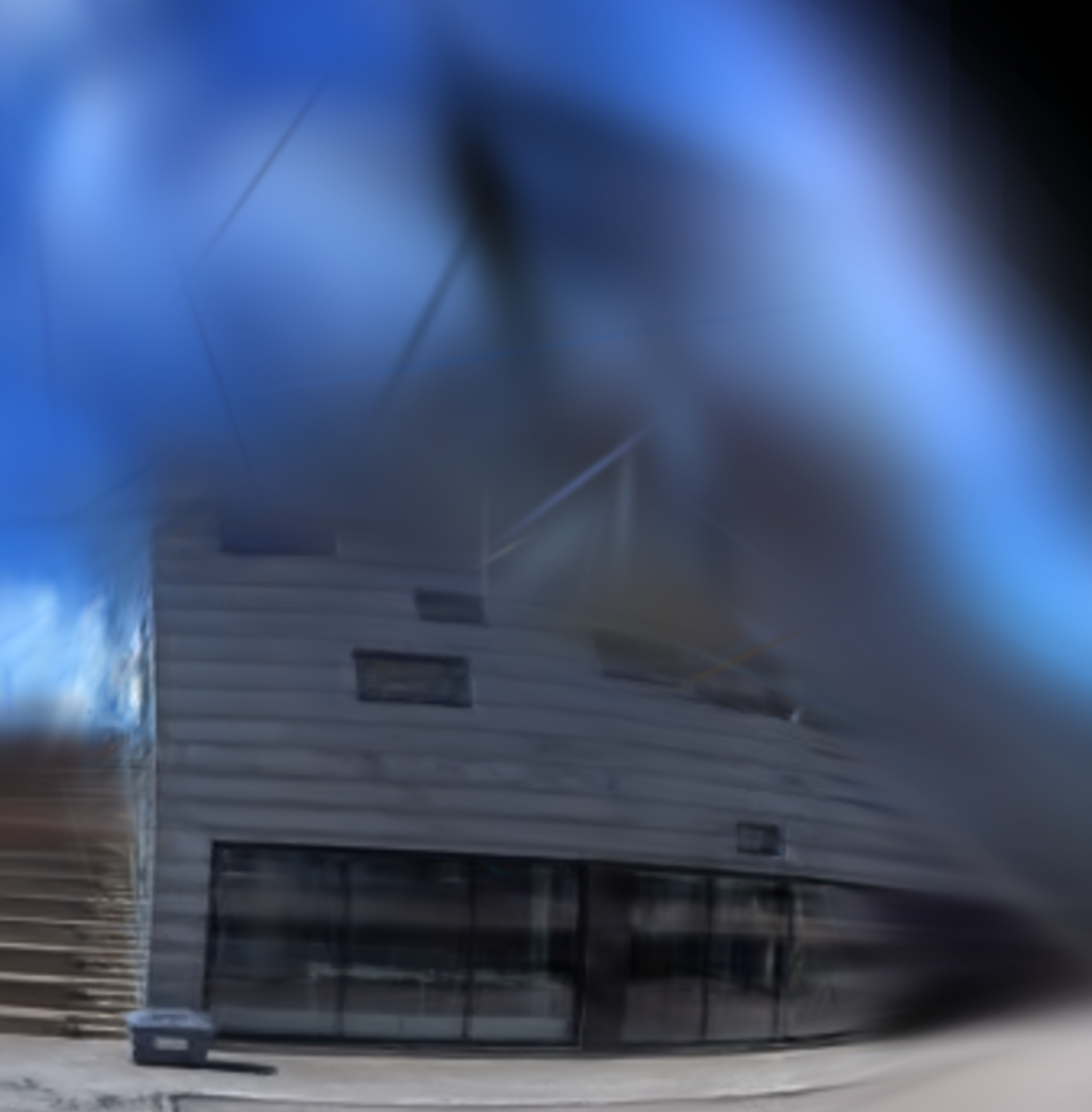}
\includegraphics[width=0.19\textwidth]{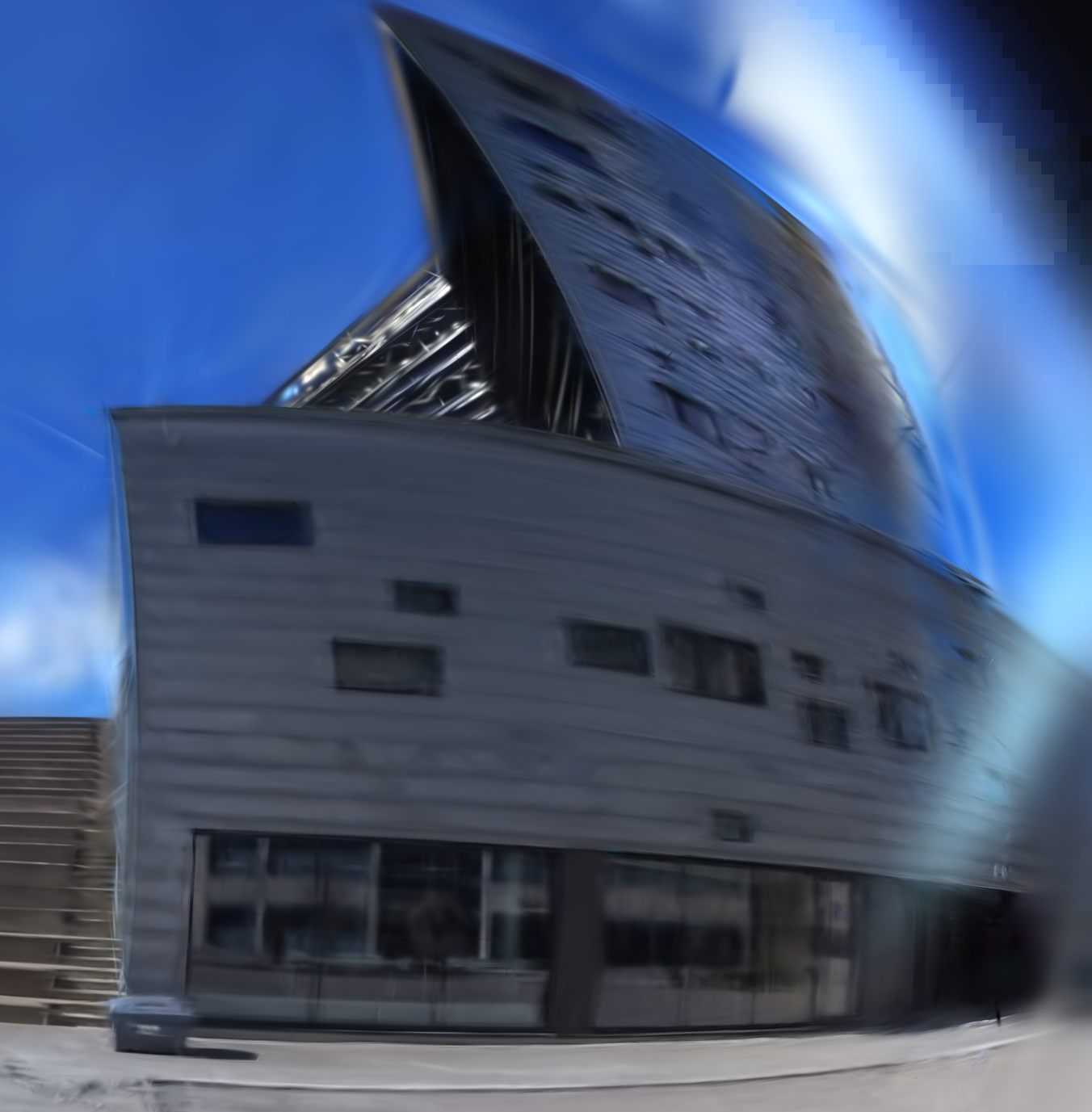}
\includegraphics[width=0.19\textwidth]{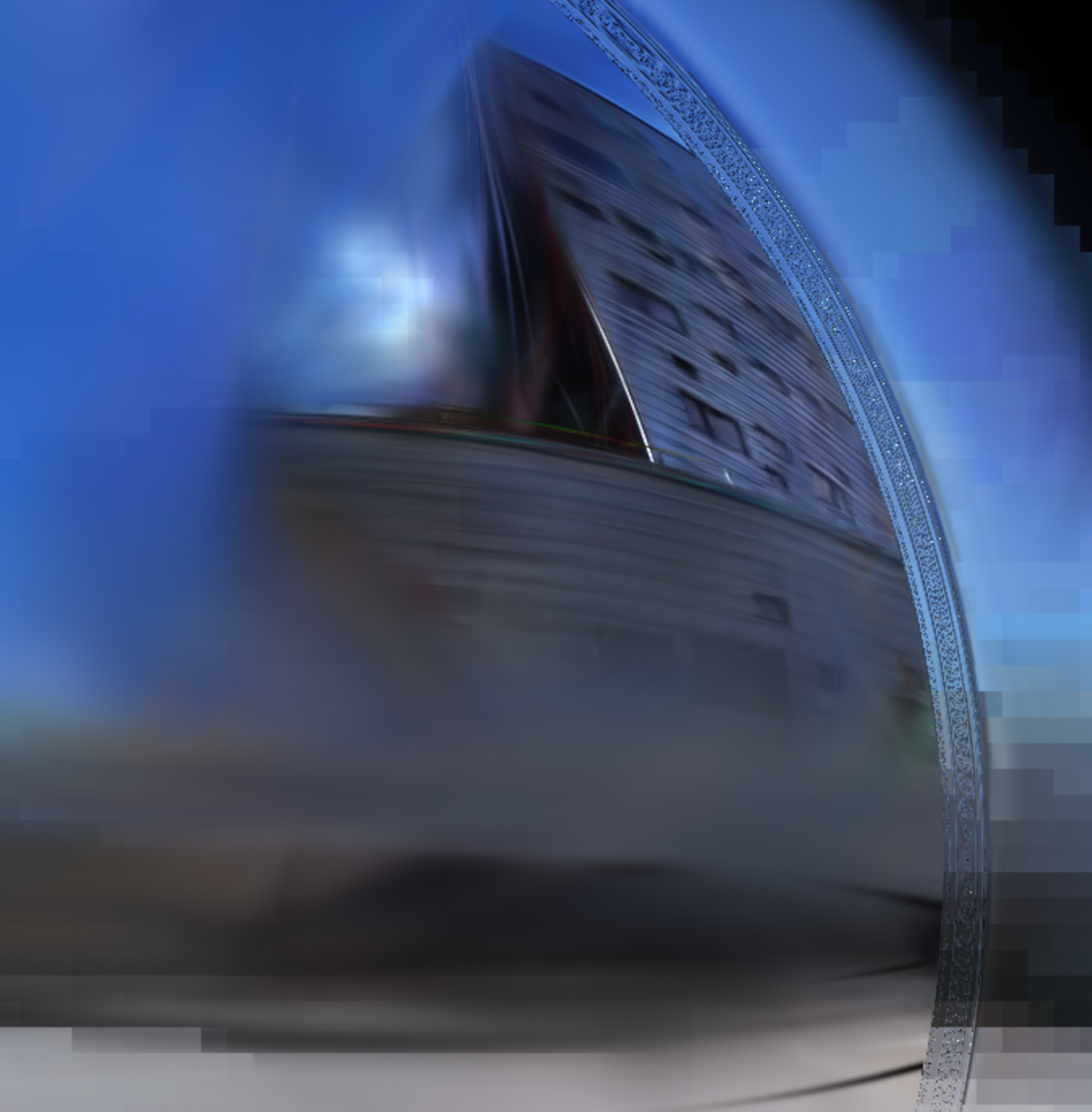}

\caption{Example renderings from Hall and Building Out using SfM- and UniK3D-based initialization (FGS: Fisheye-GS). Towards the periphery, depth-based results degrade for both methods due to projection ambiguity, while in central regions they often recover sharper details.}
\label{fig:init_strategy_comparison}
\end{figure*}
\subsection{Camera Calibration and Data}
We use the FIORD dataset \cite{gunes2025fiord}, which provides 5 indoor and 5 outdoor real-world scenes, each with 200–400 high-resolution fisheye images captured using the Insta360 One RS 1-Inch camera \cite{insta360}. Each dual-lens capture is split into two 3264×3264 fisheye PNGs with a $200^\circ$ FoV. The dataset spans a variety of scales and challenging conditions (fog, snow, glare, clutter), making it suitable for evaluating wide-FoV reconstruction.

Both lenses are calibrated independently using the Camera Calibration Toolbox for Generic Lenses \cite{1642666,kannala2008geometric}, and the resulting intrinsics and distortion parameters are converted to OpenCV’s fisheye model for use in COLMAP’s SfM pipeline. Calibration accuracy is verified by rectifying sample images and visually checking the projected geometry. FIORD is one of the few real datasets providing $200^\circ$ fisheye imagery with COLMAP-compatible calibrations, making it well suited for controlled evaluation. We use a standard 90/10 train–test split and report PSNR, SSIM, and LPIPS \cite{LPIPS} metrics on held-out views.

\subsection{Evaluated Gaussian Splatting Methods for Fisheye Cameras}

We evaluate two splatting pipelines designed for fisheye imagery: Fisheye-GS \cite{liao2024fisheyegslightweightextensiblegaussian} and 3DGUT \cite{wu20253dgut}. Both extend 3DGS \cite{kerbl3Dgaussians} to non-linear camera models. We use the official implementations and adapt them to our calibrated fisheye inputs. Splatting requires camera poses and an initial point cloud; we therefore benchmark both traditional SfM initialization and monocular depth-based initialization (UniK3D), which avoids multi-view matching and leverages the large scene coverage of a single fisheye.

\paragraph{Structure-from-Motion (SfM)}
We run COLMAP v3.9.1 \cite{schoenberger2016sfm} using the OpenCV fisheye model \cite{opencv_library}. Scenes are processed with calibrated intrinsics (fx, fy, cx, cy, k1–k4). High-resolution feature extraction and vocabulary-tree matching are enabled to improve correspondences under wide distortion. After incremental registration and triangulation, bundle adjustment refines poses and sparse geometry, which are exported for downstream splatting.

\paragraph{Fisheye-GS Implementation}
Fisheye-GS is evaluated on the full $200^\circ$ dataset, including scenes with reflections and low-texture regions where distortion is strongest. The method assumes an equidistant projection, where radial displacement scales linearly with the ray’s incidence angle. Input images are reprojected using COLMAP intrinsics, but the implementation omits the $k_1$ distortion term to maintain the distortion-free equidistant assumption. Since the equidistant model is defined only up to $180^\circ$ ($\theta=\pi$), our $200^\circ$ imagery introduces projection ambiguity near the image boundary, allowing us to assess robustness under extreme FoV.

\paragraph{3DGUT Implementation}
3DGUT is evaluated on the same $200^\circ$ fisheye data. Unlike Fisheye-GS, it directly applies the full non-linear fisheye model without reprojection and uses the Unscented Transform \cite{UT} to propagate sigma points through the projection function. This avoids local linearization and improves behavior near the periphery. We enable the optional MCMC training mode \cite{kheradmand20253dgaussiansplattingmarkov}, which updates Gaussians via Langevin dynamics \cite{brosse2018promisespitfallsstochasticgradient} and replaces heuristic densification. We use the official implementation and adapt it to the intrinsics of fisheye images.

3DGUT estimates its maximum field of view ($\theta_{\max}$) from image resolution, principal point, and focal length using a perspective-inspired approximation. While suitable for perspective datasets, this assumption can be inaccurate for ultra-wide fisheye imagery, providing an additional point of comparison in our experiments.
\textbf{\begin{table*}[htb!]
\centering\scriptsize
\renewcommand*{\arraystretch}{.8} 
\caption{Performance of the Fisheye-GS and 3DGUT methods across scenes and FOVs. SfM based initialization is used. Metrics: PSNR ↑ / SSIM ↑ / LPIPS ↓. The best average results are bolded.}
\label{tab:scene_fov_results}
\resizebox{\textwidth}{!}{%
\begin{tabular}{l|l|ccc|ccc|ccc}
\toprule
\multirow{2}{*}{\textbf{Scene}} & \multirow{2}{*}{\textbf{Method}} & \multicolumn{3}{c}{\textbf{FOV = 200°}} & \multicolumn{3}{|c}{\textbf{FOV = 160°}} & \multicolumn{3}{|c}{\textbf{FOV = 120°}} \\
\cline{3-11}
 & & PSNR$\uparrow$ & SSIM$\uparrow$ & LPIPS$\downarrow$ & PSNR$\uparrow$ & SSIM$\uparrow$ & LPIPS$\downarrow$ & PSNR$\uparrow$ & SSIM$\uparrow$& LPIPS$\downarrow$ \\
\midrule
\texttt{Kitchen} & Fisheye-GS & 20.05 & 0.7238 & 0.3862 & 24.01 & 0.8190 & 0.2509 & 19.43 & 0.7713 & 0.2569 \\
        & 3D-GUT     & 22.16 & 0.8140 & 0.3242 & 23.29 & 0.8621 & 0.2131 & 20.33 & 0.8402 & 0.2128 \\
\midrule
\texttt{Hall}    & Fisheye-GS & 19.62 & 0.7008 & 0.4024 & 24.04 & 0.8445 & 0.2203 & 20.98 & 0.8314 & 0.2075 \\
        & 3D-GUT     & 20.91 & 0.7787 & 0.3594 & 22.67 & 0.8783 & 0.1946 & 21.32 & 0.8766 & 0.1812 \\
\midrule
\texttt{Meeting} & Fisheye-GS & 17.75 & 0.6921 & 0.4540 & 20.07 &0.7468 &0.3363 & 19.19 & 0.6901 & 0.3368 \\
        & 3D-GUT     & 20.93 & 0.8171 & 0.3155 & 22.32& 0.8653 & 0.2168 & 21.00 & 0.8331 & 0.2781 \\
\midrule
\texttt{Building} & Fisheye-GS & 18.57 & 0.6505 & 0.4828 & 20.59& 0.7477 & 0.3181 & 16.21 & 0.6991 & 0.3271 \\
         & 3D-GUT     & 17.54 & 0.6281 & 0.4811 & 19.46& 0.7792 & 0.4150 & 16.99 & 0.7546 & 0.3885 \\
\midrule
\texttt{Upstairs} & Fisheye-GS & 19.42& 0.7231 & 0.4234 & 19.28 & 0.7112 & 0.3932 & 17.70 & 0.7004 & 0.3607 \\
         & 3D-GUT     & 16.64    & 0.5582     & 0.6955     & 20.61& 0.8181 & 0.4083 & 17.27 & 0.7467 & 0.6198 \\
\midrule
\texttt{Corridor} & Fisheye-GS & 19.86 & 0.6432 & 0.3648 & 20.58& 0.7332 & 0.2527 & 15.75 & 0.6394 & 0.3179 \\
         & 3D-GUT     & 20.25 & 0.6261 & 0.3385 & 20.35& 0.7422 & 0.2413 & 16.37 & 0.6641 & 0.2745 \\
\midrule
\texttt{Building Out} & Fisheye-GS & 17.11 & 0.5664 & 0.4877 & 19.20& 0.6539 & 0.3652 & 18.14 & 0.6596 &0.3411 \\
             & 3D-GUT     & 18.54& 0.6479 & 0.4461 & 18.22 & 0.7510 & 0.3002 & 17.66 & 0.7375 & 0.2286 \\
\midrule
\texttt{Night}  & Fisheye-GS & 33.18& 0.9204 & 0.1838 & 28.75 & 0.8710 & 0.2383 & 32.62 & 0.9182 & 0.1678 \\
        & 3D-GUT     & 28.19 & 0.8910 & 0.1677 & 28.57& 0.8940 & 0.2110 & 24.61 & 0.8667 & 0.2482 \\
\midrule
\texttt{Bridge}  & Fisheye-GS & 23.00 & 0.7529 & 0.3606 & 22.91 & 0.8311 & 0.2318 & 17.37 & 0.7835 & 0.2649 \\
        & 3D-GUT     & 22.06 & 0.7651 & 0.4150 & 21.39 & 0.8586 & 0.2059 & 18.21 & 0.8235 & 0.2591 \\
\midrule
\texttt{Road}    & Fisheye-GS & 18.21 & 0.6178 & 0.4552 & 19.71 & 0.7099 & 0.3357 & 15.02 & 0.6397 & 0.3746 \\
        & 3D-GUT     & 17.57    & 0.6405     & 0.4998     & 19.53 & 0.7121 & 0.3954 & 17.07 & 0.7211 & 0.3957 \\
\toprule
\textbf{Average}    & Fisheye-GS & 20.67 & 0.6991 & 0.4001 & \textbf{21.91} & 0.7668 & 0.2942 & 19.24 & 0.7333 & 0.2955 \\
& 3D-GUT     & 20.48    & 0.7167     & 0.4043     & 21.64 & \textbf{0.8161} & \textbf{0.2802} & 19.08 & 0.7864 & 0.3087 \\
\bottomrule
\end{tabular}
}
\end{table*}}

\subsection{Field of View Adjustment}
\label{sec:fov}
Our fisheye cameras capture 200\textdegree{} views, exceeding the 180\textdegree{} limit of the equidistant projection assumed by Fisheye-GS. Beyond this range, rays originate from behind the optical axis and cannot be projected consistently. While 3DGUT supports arbitrary camera models, both methods show sensitivity to distortion near the image boundary at wide angles.

To investigate the tradeoff between peripheral distortion and scene coverage, we generate additional image sets at 160\textdegree{} and 120\textdegree{}. Each 200\textdegree{} input is reprojected to the target FoV under the equidistant model by adjusting the focal length, rescaling angular coordinates, and discarding pixels beyond the desired angular range. Each FoV variant is then used to train and evaluate both Fisheye-GS and 3DGUT to assess how reconstruction quality changes under reduced angular input. 


    



\subsection{Gaussian Initialization: SfM vs UniK3D}
SfM pipelines often degrade on fisheye images due to distortion and sparse feature matches. As an alternative, we initialize Gaussian Splatting using monocular 3D estimates from UniK3D \cite{piccinelli2025unik3d}, which supports arbitrary intrinsics—including fisheyes, unlike most trained depth estimators. This makes UniK3D well suited for our wide-FoV setting, where standard monocular models fail to produce consistent geometry.

We use only 2 fisheye views per scene, leveraging the wide FoV to capture most of the environment in a few frames while avoiding inconsistencies that arise when fusing many monocular predictions. Despite the minimal input, the fused point clouds provide sufficient structure for Gaussian initialization.

UniK3D outputs points in the local camera frame. We convert them to the COLMAP world frame using estimated extrinsics, fuse points across the selected views, and apply a similarity transform to match COLMAP’s scale. Because UniK3D produces dense geometry, we downsample its point clouds using voxel-based sampling to achieve approximate parity with SfM point counts while preserving spatial coverage. Exact matching is impossible without altering structure, but approximate equality is sufficient for a fair comparison. The number of initialization points for both methods is reported in Table~\ref{tab:sfm_depth_by_method_headers}.

To verify alignment quality, we compute 2D–3D correspondences between UniK3D predictions and SfM points and report reprojection errors as a quantitative check. This confirms that UniK3D’s point clouds are sufficiently consistent with COLMAP for use in both Fisheye-GS and 3DGUT. Although UniK3D has not been trained on real fisheye images with extreme FoVs (e.g., $200^\circ$), our setup allows us to directly evaluate its suitability as a geometry source for 3D Gaussian initialization.
\begin{table*}[ht]
\caption{Performance comparison of 3D Gaussian splatting renderings for SfM based and UniK3D based Gaussian initialization across scenes at FOV=200\textdegree{}. Rows stack SfM (top) over Depth (bottom). Metrics: PSNR~$\uparrow$ / SSIM~$\uparrow$ / LPIPS~$\downarrow$. Init Pts refers to the number of initial Gaussian points. The best average results are bolded.}
\label{tab:sfm_depth_by_method_headers}
\centering\scriptsize
\setlength{\tabcolsep}{9pt} 
\renewcommand{\arraystretch}{0.8}
\begin{tabular}{l|l|c|ccc|ccc}
\toprule
\multirow{2}{*}{\textbf{Scene}} & \multirow{2}{*}{\textbf{Init}} & 
\multirow{2}{*}{\textbf{\# Init Pts}} &
\multicolumn{3}{c}{\textbf{Fisheye-GS}} &
\multicolumn{3}{|c}{\textbf{3D-GUT}} \\
\cline{4-9}
& & & PSNR$\uparrow$ & SSIM$\uparrow$ & LPIPS$\downarrow$ & PSNR$\uparrow$ & SSIM$\uparrow$ & LPIPS$\downarrow$ \\
\midrule
\multirow{2}{*}{\texttt{Kitchen}}
& SfM   & 82K  & 20.05 & 0.7238 & 0.3862 & 22.16 & 0.8140 & 0.3242 \\
& Depth & 85K  & 21.80 & 0.7751 & 0.2982 & 20.63 & 0.7782 & 0.4147 \\
\midrule
\multirow{2}{*}{\texttt{Hall}}
& SfM   & 147K & 19.62 & 0.7008 & 0.4024 & 20.91 & 0.7787 & 0.3594 \\
& Depth & 149K & 22.16 & 0.7824 & 0.3010 & 19.80 & 0.7732 & 0.4371 \\
\midrule
\multirow{2}{*}{\texttt{Meeting}}
& SfM   & 55K  & 17.75 & 0.6921 & 0.4540 & 20.93 & 0.8171 & 0.3155 \\
& Depth & 37K  & 17.63 & 0.6692 & 0.4648 & 20.48 & 0.8165 & 0.4009 \\
\midrule
\multirow{2}{*}{\texttt{Building}}
& SfM   & 194K & 18.57 & 0.6505 & 0.4828 & 17.54 & 0.6281 & 0.4811 \\
& Depth & 189K & 19.86 & 0.6647 & 0.4033 & 18.62 & 0.7230 & 0.4480 \\
\midrule
\multirow{2}{*}{\texttt{Upstairs}}
& SfM   & 186K & 19.42 & 0.7231 & 0.4234 & 16.64 & 0.5582 & 0.6955 \\
& Depth & 195K & 20.95 & 0.7387 & 0.3568 & 14.68 & 0.5286 & 0.7222 \\
\midrule
\multirow{2}{*}{\texttt{Corridor}}
& SfM   & 219K & 19.86 & 0.6432 & 0.3648 & 20.25 & 0.6261 & 0.3385 \\
& Depth & 221K & 21.90 & 0.7263 & 0.3072 & 20.49 & 0.6244 & 0.3951 \\
\midrule
\multirow{2}{*}{\texttt{Building Out}}
& SfM   & 140K & 17.11 & 0.5664 & 0.4877 & 18.54 & 0.6479 & 0.4461 \\
& Depth & 150K & 18.31 & 0.6125 & 0.3940 & 18.46 & 0.6430 & 0.5140 \\
\midrule
\multirow{2}{*}{\texttt{Night}}
& SfM   & 20K  & 33.18 & 0.9204 & 0.1838 & 28.19 & 0.8910 & 0.1677 \\
& Depth & 22K  & 35.98 & 0.9427 & 0.1177 & 29.00 & 0.8960 & 0.2341 \\
\midrule
\multirow{2}{*}{\texttt{Bridge}}
& SfM   & 235K & 23.00 & 0.7529 & 0.3606 & 22.06 & 0.7651 & 0.4150 \\
& Depth & 238K & 24.87 & 0.7901 & 0.2935 & 22.02 & 0.7624 & 0.4203 \\
\midrule
\multirow{2}{*}{\texttt{Road}}
& SfM   & 352K & 18.21 & 0.6178 & 0.4552 & 17.57 & 0.6405 & 0.4998 \\
& Depth & 352K & 18.36 & 0.6244 & 0.4440 & 16.12 & 0.5747 & 0.6081 \\
\midrule
\multirow{2}{*}{\textbf{Average}}
& SfM   & 163K & 20.68 & 0.6991  & 0.4001  & 20.48 & 0.7167 & 0.4043 \\
& Depth & 164K & \textbf{22.18} & \textbf{0.7326}  & \textbf{0.3381}  & 20.03 & 0.7120 & 0.4595  \\
\bottomrule
\end{tabular}
\end{table*}
\section{\uppercase{Results}}
\subsection{Evaluation on Real Fisheye Images}

Our primary evaluation uses the full $200^\circ$ fisheye inputs, where distortion is strongest. Table~\ref{tab:scene_fov_results} reports quantitative results across ten FIORD scenes, spanning compact indoor spaces (Kitchen, Hall, Meeting), larger indoor areas (Building, Upstairs), small outdoor regions (Corridor, Night, Building Out), and large outdoor environments (Bridge, Road). This diversity allows us to analyze how scene scale and capture extent influence reconstruction quality.

Reconstruction quality depends strongly on scene scale. In compact indoor scenes, 3DGUT clearly outperforms Fisheye-GS, showing PSNR gains of 2–3 dB and consistently higher SSIM and lower LPIPS. This reflects its strength in modeling non-linear distortion when the scene is bounded or semi-bounded. In contrast, in larger indoor and outdoor environments, 3DGUT loses this advantage and sometimes falls behind Fisheye-GS. These scenes involve longer baselines, wider spatial coverage, and more input views, which amplify the limitations of 3DGUT’s FoV approximation. At $200^\circ$, 3DGUT often produces blurred reconstructions at the periphery. This behavior aligns with its projection formulation, which estimates the maximum field of view using a perspective-inspired approximation that becomes inaccurate at ultra-wide angles, leading to projection ambiguity near the $200^\circ$ boundary. Outdoor scenes further challenge the method due to low texture (sky, snow, repetitive surfaces) and strong lighting variation (glare, overexposure, nighttime), causing noticeable degradation in the largest scenes (Upstairs, Road).

Overall, 3DGUT preserves peripheral detail well in compact scenes but becomes unstable in larger-scale settings. Fisheye-GS, despite its simplified equidistant model, offers more stable performance in these cases due to its tight coupling with COLMAP’s sparse initialization. These results match the intended behavior of the two methods: 3DGUT performs best in small distortion-heavy scenes, while Fisheye-GS trades local fidelity for greater robustness in large multi-view environments. Our experiments provide the first empirical confirmation of these theoretical trade-offs on real fisheye imagery at $200^\circ$ FoV.

\subsection{Impact of Field of View Adjustment}

We evaluate both methods under reduced FoVs of $160^\circ$ and $120^\circ$ to study the trade-off between peripheral distortion and scene coverage while testing how these two elements affect the rendering results. As shown in Table~\ref{tab:scene_fov_results}, $160^\circ$ FoV consistently provides the best compromise.

For Fisheye-GS, trimming the FoV from $200^\circ$ to $160^\circ$ yields clear improvements in SSIM and LPIPS across most scenes (Kitchen, Hall, Meeting, Building), confirming its sensitivity to peripheral distortion. Even when PSNR or SSIM remain slightly higher at $200^\circ$, perceptual quality still improves at $160^\circ$ due to lower LPIPS. The only exception is the Night scene, where $200^\circ$ shows marginally higher PSNR/SSIM because dark peripheral regions inflate pixel-wise scores after trimming. This does not translate into perceptual gains, and visual quality remains comparable. At $120^\circ$, performance generally drops again: distortion is reduced, but excessive cropping removes important scene content, lowering PSNR and reducing contextual consistency.

3DGUT shows relatively stable PSNR across FoVs, but SSIM and LPIPS consistently improve at $160^\circ$, indicating perceptually closer reconstructions even when pixel-wise differences remain similar. This suggests that $160^\circ$ FoV reduces distortion effects without sacrificing necessary coverage. Overall, $160^\circ$ provides the most balanced FoV for both methods: wide enough to maintain scene context, while sufficiently narrow to limit peripheral distortion.

\subsection{Impact of Depth-Based Initialization}
\label{sec:depth_init_results}
Table~\ref{tab:sfm_depth_by_method_headers} reports results using monocular depth-based initialization versus SfM for Fisheye-GS and 3DGUT. To avoid bias from denser UniK3D monocular predictions, we applied uniform sampling to the UniK3D outputs so that the point counts were comparable to the sparse SfM reconstructions. This ensures that differences in the performances obtained by our two methods are attributable to initialization quality (geometric accuracy of the scene) rather than point count.

For Fisheye-GS, depth-based initialization performs competitively across scenes: most metrics remain close to SfM, with slight improvements in some cases (e.g., Kitchen, Hall, Upstairs, Bridge). The main benefit is more accurate geometric coverage from fewer input views, obtained in a fraction of the time: UniK3D produces usable geometry in about 10 seconds, compared to roughly an hour for a full SfM reconstruction. This faster initialization helps capture fine structure reducing preprocessing cost.

For 3DGUT, depth-based initialization generally yields lower performance than SfM in several scenes (e.g., Upstairs, Road), with only mixed gains in others (e.g., Meeting, Building Out). The largest performance drops occur near the periphery, where strong distortion and projection ambiguity at ultra-wide FoVs amplify errors. Figure \ref{fig:init_strategy_comparison} illustrates this trend: depth initialization degrades peripheral regions due to projection ambiguity, while central regions may even achieve sharper detail. This spatial trade-off explains the overall metric drop for 3DGUT, despite localized gains in accuracy.

Taken together, these findings show that depth-based initialization is a practical substitute for SfM in fisheye Gaussian Splatting. While its benefits at 200\textdegree{} are muted by the limitations of 3DGUT’s projection, the approach offers substantial savings in preprocessing time and remains effective in distortion-heavy or challenging indoor scenes where SfM often generates floaters or is costly to compute.
\section{Conclusion}
We presented the first systematic evaluation of fisheye-adapted 3D Gaussian Splatting on real imagery with fields of view beyond $180^\circ$. Across the FIORD dataset, 3DGUT showed advantages in compact environments due to its non-linear distortion handling, while Fisheye-GS proved more stable in large scenes through its simplified distortion model and SfM coupling. FoV reduction experiments reinforced these trends: narrowing the FoV reduces peripheral distortion, with $160^\circ$ offering a strong balance between coverage and geometric stability, whereas $120^\circ$ removes too much context. We also evaluated depth-based initialization using UniK3D. Despite lacking training on real fisheye images, UniK3D generalized well and enabled Fisheye-GS reconstructions comparable to SfM while reducing preprocessing time substantially. However, for 3DGUT, peripheral distortion limited reconstruction quality.

Overall, our results show that fisheye Gaussian Splatting is feasible without heavy preprocessing and that monocular depth can serve as a practical alternative to SfM in distortion-heavy scenes. Future work may explore regressing Gaussian parameters directly from monocular 3D estimators and extending these approaches to larger-scale scenes.
\subsection*{ACKNOWLEDGEMENTS}
We acknowledge the financial support of the Intelligent Work Machines Doctoral Education Pilot Program (IWM VN/3137/2024-OKM-4) and the Academy of Finland projects 353139 and 362409. We also acknowledge CSC – IT Center for Science, Finland, for computational resources.
\bibliographystyle{apalike}
{\small
\bibliography{example.bib}}
\section*{\uppercase{Appendix}}
\subsection*{Preliminaries}
\label{sec:preliminaries}
\paragraph{Fisheye-GS} adapts the 3D Gaussian Splatting pipeline to operate directly on distorted fisheye imagery by replacing perspective projection with an equidistant model. In this formulation, the radial distance from the optical center to the projected point on the image plane is proportional to the angle of incidence \(\theta\) between the incoming ray and the optical axis:
\begin{equation}
    r = f \cdot \theta
\end{equation}
where \(\theta\) is defined for a 3D point \(\mathbf{p}_c = (x_c, y_c, z_c)^\top\) in camera coordinates as:
\begin{equation}
    \theta = \arctan\left( \frac{\sqrt{x_c^2 + y_c^2}}{z_c} \right)
\end{equation}
To support gradient-based optimization, Fisheye-GS derives the Jacobian of its projection function and integrates it into the rasterization backend. This enables accurate backpropagation while preserving spatial fidelity in wide-angle views. Aside from the projection module, the remaining pipeline—Gaussian initialization, filtering, and rendering—follows the original 3DGS implementation.

Although the equidistant model provides a reasonable approximation up to $\theta=\pi$ ($180^\circ$), it becomes inaccurate when the field of view exceeds this range. In our setup, the cameras capture up to $200^\circ$, meaning rays with $\theta>\pi$ originate from behind the optical axis. Such rays cannot be consistently projected onto a forward-facing image plane, leading to unavoidable errors near the image periphery.
\paragraph{3D Gaussian Unscented Transform (3DGUT)} takes a different approach to handling non-linear camera models by replacing the Elliptical Weighted Average (EWA) splatting \cite{zwicker2002ewa} used in standard 3DGS with the Unscented Transform (UT) \cite{UT}. Rather than projecting a single Gaussian mean, each 3D Gaussian is represented as a set of deterministically sampled sigma points that approximate the original distribution:
\begin{equation}
    \{\mathbf{x}_i\} = \text{UT}(\boldsymbol{\mu}, \Sigma)
\end{equation}
 Each sigma point \(\mathbf{x}_i\) is projected individually through the camera model, which avoids local linearization and captures non-linear effects such as fisheye distortion more faithfully. This results in a more accurate rendering of Gaussian contributions, particularly near the image periphery where standard approximations break down.

\end{document}